\documentclass{article}

\usepackage{arxiv}
\usepackage{natbib}

\usepackage[utf8]{inputenc} 
\usepackage[T1]{fontenc}    
\usepackage{hyperref}       
\usepackage{url}            
\usepackage{booktabs}       
\usepackage{amsfonts}       
\usepackage{nicefrac}       
\usepackage{microtype}      
\usepackage{lipsum}
\usepackage{graphicx}
\usepackage{booktabs} 
\usepackage{epsfig}
\usepackage{algorithm}  
\usepackage{subcaption}
\usepackage{graphicx}
\usepackage{algorithm}
\usepackage{algpseudocode}

\usepackage{algpseudocode}
\usepackage{mathtools,array}
\algtext*{EndWhile}
\algtext*{EndIf}
\algtext*{EndFor}
\algtext*{EndProcedure}
\algtext*{EndFunction}
\usepackage[marginal]{footmisc}
\usepackage{soul,color}
\usepackage{bbding}
\usepackage{adjustbox}
\usepackage{enumerate}
\usepackage{booktabs}
\usepackage{tablefootnote}
\usepackage{pifont}
\usepackage{multirow}
\usepackage[dash,dot]{dashundergaps}
\usepackage{amsmath}

\newcommand{\eat}[1]{}
\usepackage[normalem]{ulem}
\usepackage{soul}
\usepackage{float}
\usepackage{tabularx} 
\usepackage{enumitem}
\usepackage{makecell}
\usepackage{hyperref}  
\usepackage{nicefrac}  
\usepackage[justification=raggedright,singlelinecheck=false]{caption}
\graphicspath{ {./images/} }

\title{Knowledge-Guided Adaptive Mixture of Experts for Precipitation Prediction}

\author{
 Chen Jiang \\
  Samuel Ginn College of Engineering\\
  Auburn University\\
  Auburn, AL 36849 \\
  \texttt{czj0042@auburn.edu} \\
   \And
 Kofi Osei \\
  Gustavus Adolphus College\\
  St. Peter, MN, 56082 \\
  \texttt{okosei@gustavus.edu} \\
  \And
 Sai Deepthi Yeddula \\
  School of Engineering and Computer Science\\
  Oakland University\\
  Rochester, MI, 48309 \\
  \texttt{syeddula@oakland.edu} \\
    \And
 Dongji Feng \\
  School of Computing and Design\\
  California State University Monterey Bay,\\
  Seaside, CA, 93955 \\
  \texttt{dfeng@csumb.edu} \\
    \And
 Wei-Shinn Ku \\
  Samuel Ginn College of Engineering\\
  Auburn University\\
  Auburn, AL 36849 \\
  \texttt{wzk0004@auburn.edu} \\
}

\begin{document}
\maketitle
\begin{abstract}
Accurate precipitation forecasting is indispensable in agriculture, disaster management, and sustainable strategies. However, predicting rainfall has been challenging due to the complexity of climate systems and the heterogeneous nature of multi-source observational data, including radar, satellite imagery, and surface-level measurements. The multi-source data vary in spatial and temporal resolution, and they carry domain-specific features, making it challenging for effective integration in conventional deep learning models.\\

Previous research has explored various machine learning techniques for weather prediction; however, most struggle with the integration of data with heterogeneous modalities. To address these limitations, we propose an Adaptive Mixture of Experts (MoE) model tailored for precipitation rate prediction. Each expert within the model specializes in a specific modality or spatio-temporal pattern. We also incorporated a dynamic router that learns to assign inputs to the most relevant experts. Our results show that this modular design enhances predictive accuracy and interpretability.\\

In addition to the modeling framework, we introduced an interactive web-based visualization tool that enables users to intuitively explore historical weather patterns over time and space. The tool was designed to support decision-making for stakeholders in climate-sensitive sectors. We evaluated our approach using a curated multimodal climate dataset capturing real-world conditions during Hurricane Ian in 2022. The benchmark results show that the Adaptive MoE significantly outperformed all the baselines. 
\end{abstract}


\section{Introduction}
Accurate precipitation forecasting is essential for informing agriculture and disaster management. Modern weather forecasting systems have increasingly relied on integrating diverse and multi-source observational data, such as radar reflectivity, satellite imagery, and atmospheric measurements. The multi-source data adds invaluable insights into evolving weather patterns and opens more avenues for real-time monitoring and management. The tradeoff of enriched information in modeling using multi-source data is the highly heterogeneous nature of such data. For example, data coming from multiple sources can be highly diverse and variable in spatial and temporal resolution, and physical features. This is the main reason that integrating multi-source data into traditional deep learning models could be less successful compared to using single-source data.

Although recent advances in machine learning have improved weather prediction capabilities~\cite{espeholt2022deep, salman2015weather, ham2019deep}, many models remain limited in their ability to effectively process heterogeneous, multimodal data inputs. In particular, variations in data format, temporal alignment, and regional relevance complicate joint learning across modalities. Standard architecture in the most commonly used modeling methodologies often treats all inputs uniformly, neglecting the fact that different data sources may contribute differently depending on the spatial or temporal context.

To address this limitation, we introduced an adaptive MoE model designed for precipitation rate prediction. Each expert model specializes in a specific data modality or spatio-temporal segment. A dynamic router was also used to assign inputs to the most relevant experts. This modular design enables the framework to maintain a balance between specialization and scalability, leading to great enhancement in both accuracy and interpretability across complex climate conditions.

In addition, we presented a lightweight, browser-based visualization platform that enables users to explore input features in an interactive manner. This tool provides transparent insight into weather patterns across space and time.

\subsection*{Our contributions are:}
\begin{enumerate}
    \item \textbf{MoE-Climate Dataset:}  
    We introduced a curated multi-modal climate dataset tailored for multimodal precipitation forecasting tasks involving heterogeneous inputs. This dataset has been publicly released to encourage reproducibility and further research.

    \item \textbf{Pretrained MoE Model:}  
    We presented a modular model utilizing an MoE model that can adaptively integrate multiple diverse climate data sources, thus enhancing prediction accuracy.

    \item \textbf{Web-Based Interactive Interface:}  
    We developed a lightweight browser-based application that allows users to explore historical climate patterns at multiple spatial and temporal scales over a specified time period.
\end{enumerate}

This work advances the field by demonstrating how modular deep learning architectures can be effectively applied to spatio-temporal prediction for climate management and provide practical tools that support informed, data-driven decisions in high-impact environmental applications.

\section{Related Work}
\label{gen_inst}
\subsection{Spatial-Temporal Prediction}

Spatial-temporal prediction is crucial for tasks requiring accurate forecasting of evolving processes across both space and time. Various methods have emerged to solve these complex forecasting problems effectively. Neural Networks such as Recurrent Neural Networks (RNNs), Long Short-Term Memory (LSTM) networks~\cite{hochreiter1997long}, have demonstrated substantial capability in modeling temporal dependencies due to their ability to capture long-range sequential information~\cite{mcnally2018predicting}.

Convolutional Neural Networks (CNNs)~\cite{lecun2015deep} and their temporal variant, Temporal Convolutional Networks (TCNs)~\cite{lea2016temporal, hewage2020temporal}, represent another powerful alternative. TCNs offer computational advantages through parallel processing and have shown superior performance in sequence prediction tasks, including traffic forecasting and other temporal applications~\cite{yeddula2023traffic, borovykh2017conditional,jiang2024convolutional}.

In recent years, Transformers and their variants have become prominent in spatial-temporal forecasting. Transformers leverage self-attention mechanisms to effectively model long-range temporal interactions without recurrence, achieving state-of-the-art results in domains such as energy consumption forecasting, financial time series prediction, and agricultural yield prediction~\cite{wu2020adversarial, yoo2021accurate, liu2022rice}. Furthermore, hybrid models combining Transformer architectures with spatial attention and graph-based methods have emerged, significantly enhancing the capability to model complex spatial dependencies and interactions in traffic prediction and urban dynamics~\cite{lin2020self, wang2020traffic, guo2019attention}.

\subsection{Climate Prediction}

Accurate climate prediction, particularly in forecasting heavy rainfall, traditionally relies on expert meteorologists manually extracting and interpreting features from extensive atmospheric data, such as temperature, atmospheric pressure, humidity, and wind velocity. While human expertise is crucial for comprehending complex weather systems, manual methods are inherently limited by scalability and prone to errors, particularly given the increasing scale and complexity of available meteorological datasets.

Recent advancements in deep learning have significantly improved the accuracy of precipitation prediction. Convolutional-based architectures like ConvLSTM~\cite{lin2020self, kim2017deeprain} and U-Net~\cite{ronneberger2015u, trebing2021smaat} have demonstrated effectiveness in capturing both spatial and temporal precipitation patterns, surpassing traditional statistical models~\citep{shi2017deep, ayzel2020rainnet}. Moreover, integrating data from diverse real-world data sources, such as satellite imagery, radar data, and ground-based sensor measurements, has further elevated the precision of these predictive models~\citep{bi2023accurate, ravuri2021skilful, jiang2024deep}.

Advanced architectures such as the MoE model~\cite{masoudnia2014mixture}, which effectively handles multi-modal fusion of climate data, present significant promise. The MoE model facilitates specialized modeling of distinct climate conditions and data modalities, yielding robust and accurate predictions across various meteorological scenarios~\citep{dryden2022spatial, moraux2021deep}. Recent trends also highlight the incorporation of physics-informed neural networks (PINNs)~\cite{cai2021physics}, which embed physical laws directly into model architectures, enhancing both the interpretability and predictive accuracy of climate forecasting models~\citep{karniadakis2021physics, willard2022integrating}.

\section{Data}
\begin{figure}[!t]
    \centering
    \resizebox{.65\textwidth}{!}{
\includegraphics{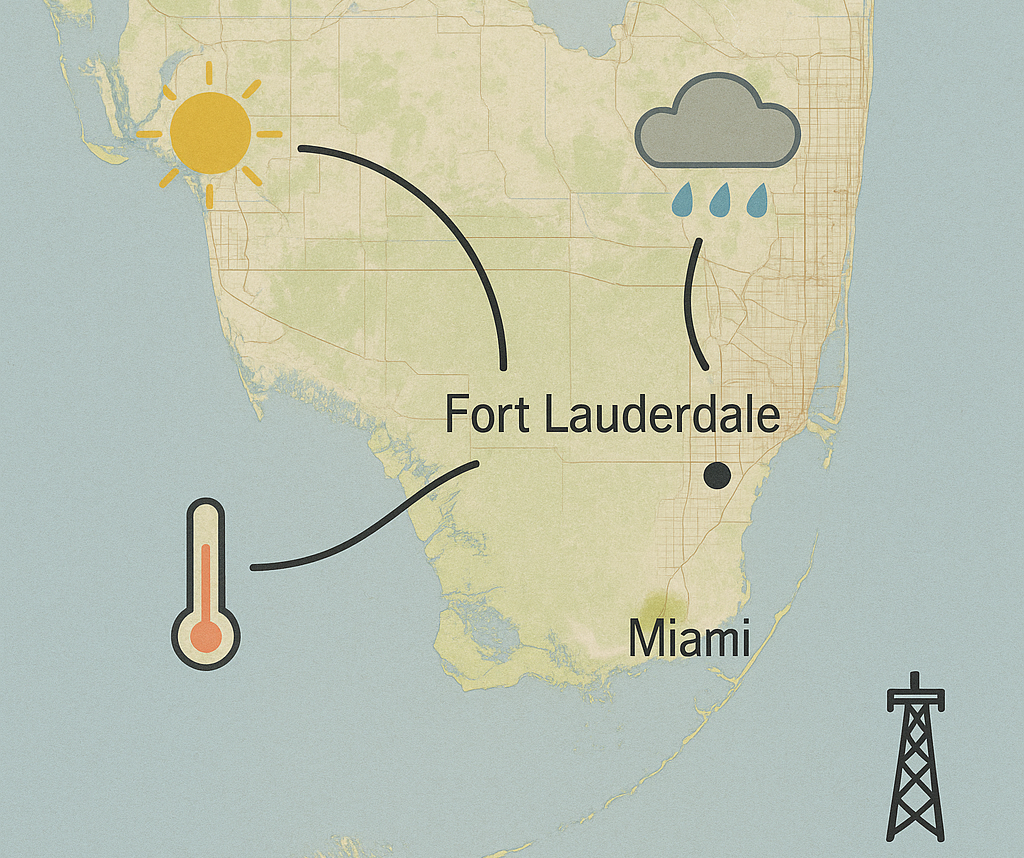}
    }
    \caption{Area of interest. The area captured in the current data set consisted of 10,000 grids across South Florida in the United States.}
    \label{fig:overview}
\end{figure}
\subsection{MoE-Climate Dataset}
This section introduces the proposed \textit{MoE-Climate} dataset, which is specifically designed to support a variety of climate-related modeling tasks using the MoE framework. The dataset serves as a valuable benchmark for exploring multimodal integration, spatial specialization, and expert routing strategies in climate modeling.

The raw data used to construct the MoE-Climate dataset was collected by the National Oceanic and Atmospheric Administration (NOAA)\footnote{\url{https://www.noaa.gov}}, leveraging multiple remote sensing platforms and in-situ devices. These sources vary in both spatial and temporal resolution, collectively covering the entire continental United States. For this study, we extracted and processed a regional subset focusing on South Florida to capture the atmospheric dynamics surrounding a specific extreme weather event.

As shown in Table~\ref{tbl:data_format}, the dataset comprises high-resolution gridded climate observations for the South Florida region during the landfall of Hurricane Ian in 2022~\cite{jiang2023multimodal}. The temporal span extends from 00:00 on September 23 to 23:00 on October 1, with an hourly resolution. Spatially, the data is organized into a $100 \times 100$ grid, where each grid cell corresponds to a 3 km $\times$ 3 km area, allowing for fine-grained monitoring of localized weather patterns.

Each grid point includes a comprehensive set of atmospheric variables sampled across 50 vertical pressure levels. These variables include precipitation rate, total accumulated precipitation, cloud cover, and 18 additional climate-related features. As detailed in Table~\ref{tbl:data_format}, the dataset also provides metadata such as grid ID, latitude, longitude, global grid dimensions (1799 $\times$ 1059), and grid spacing. Each observation is timestamped, allowing for dynamic temporal analysis throughout the entire hurricane event.

\subsection{Data Organization}
\textbf{Temporal climate data} (\(x_t\)): This captures the historical evolution of climate variables over time. To represent temporal patterns, we define \(x_t\) as the climate feature of interest (e.g., temporal precipitation rate or total accumulated precipitation) over the past 6 hours. This is represented as a sequence of \textit{N} timestamps:
\begin{equation}
x_t = \{x_t^1, x_t^2, \ldots, x_t^N\}
\end{equation}
where \(x_t^i\) (for \(i \in \{1, \ldots, N\}\)) denotes the average feature value for the \(i\)-th timestamp.

\subsection{Knowledge-Guided Feature Grouping}
To support the design of our modular MoE model, we organized the full set of climate variables into subsets based on shared physical properties. This knowledge-guided partitioning is informed by domain-specific classifications from the National Oceanic and Atmospheric Administration (NOAA)~\cite{ncei2024}. As shown in Table~\ref{tbl:data_grouping}, the features are grouped into six physically meaningful categories: Momentum, Temperature, Moisture, Mass, Cloud, and Radiation. This structure not only enhances model interpretability but also allows each expert in the MoE to specialize in a coherent subset of features.

For context, Table~\ref{tbl:dataset_comparison} presents a comparative overview of our curated MoE precipitation dataset alongside other commonly used gridded climate datasets, highlighting differences in feature dimensionality, spatial resolution, and temporal frequency.

\begin{table}[!h]
    \caption{Comparisons of climate datasets.}
    \label{tbl:dataset_comparison}
    \begin{adjustbox}{width=0.85\textwidth}
    \begin{tabular}{l | c c c c}
    \hline
     Datasets  & \ Number of Features &\ Temporal Resolution &\ Spatial Resolution\\
     \hline
      PRISM M2~\cite{daly1994statistical, daly2008physiographically}& 12 & Daily and Monthly & 4km and 6km \\
     
      H10~\cite{hamlet2010final} & 21 & Daily & 6km \\
      
      L15~\cite{livneh2015spatially}& 15 & Daily & 6km \\
      
      NLDAS~\cite{cosgrove2003real} & 15 & Hourly & 12km \\
     
      Daymet~\cite{thornton1997generating} & 7 & Daily & 1km \\
      
      N15~\cite{newman2015gridded} & 2 & Daily & 12km \\
     
      MoE Dataset (\textit{Ours})   & 19 & Hourly &3km \\
    \hline
    \end{tabular}
    \end{adjustbox}
    \caption*{PRISM: Parameter-elevation Regressions on Independent Slopes Model; NLDAS:North American Land Data Assimilation System; Daymet:Daily Surface Weather and Climatological Summaries on a 1-km Grid for North America.}
    \label{tab:data_cmp1}
\end{table}

\begin{table*}[!h]
\caption{A snapshot of the climate data used in the study that contained both spatial and temporal inputs. The dataset included grid metadata as well as time-series measurements such as precipitation rate, total precipitation, cloud cover, and 16 additional meteorological features.}
\label{tbl:data_format}
\begin{adjustbox}{width=0.85\textwidth}
\begin{tabular}{ c|c c c c c c c c}

\toprule
\textbf{GridID} & \multicolumn{1}{c|}{\textbf{Longitude}} & \multicolumn{1}{c|}{\textbf{Latitude}} & \multicolumn{1}{c|}{\textbf{Grid Points}}& \multicolumn{1}{c|}{\textbf{Grid Spacing}} & \multicolumn{1}{c}{\textbf{Vertical Level}}\\
\midrule
\multicolumn{1}{c|}{1}& \multicolumn{1}{c|}{122.71} & \multicolumn{1}{c|}{21.13} & \multicolumn{1}{c|}{1799 $\times$ 1059} & \multicolumn{1}{c|}{3 km} & \multicolumn{1}{c}{50}\\
 \bottomrule
 \toprule
\textbf{Time Stamp}& \textbf{2022/09/23 00:00} & \textbf{2022/09/23 01:00} &  \textbf{2022/09/23 02:00} & ... & \textbf{2022/10/02 00:00} \\
\midrule
Precipitation rate (mm/hour) &0.0& 0.72 & 0.94& ...&0\\
Total Precipitation (mm) &0.01& 1.88 & 4.3& ...&31.61\\
... &...& ... & ...& ...&...\\
Cloud cover &50\%& 70\% & 100\%& ...&20\%\\
\bottomrule
\end{tabular}
\end{adjustbox}
\label{tbl:data_format2}
\end{table*}
\label{headings}

\begin{table*}[!h]
\centering
    \caption{Feature group mapping. Specific features are referred to as their corresponding Feature ID in the released source code.}
    \label{tbl:data_grouping}
    \begin{tabular}{llc}
        \toprule
        \textbf{Feature Group} & \textbf{Features} & \textbf{Feature ID} \\
        \midrule
        Momentum & Wind speed & 5 \\
                 & U component of wind & 8 \\
                 & V component of wind & 9 \\
        \midrule
        Temperature & 2 metre temperature & 6 \\
                    & Dew point temperature & 10 \\
        \midrule
        Moisture & Precipitation rate & 1 \\
                 & Plant canopy surface water & 3 \\
                 & Moisture availability & 11 \\
                 & Total precipitation & 12 \\
                 & Relative humidity & 16 \\
        \midrule
        Mass & Pressure:CloudBase & 7 \\
             & Pressure:CloudTop & 15 \\
        \midrule
        Cloud & Cloud cover & 2 \\
              & Low cloud cover & 13 \\
              & Medium cloud cover & 14 \\
        \midrule
        Radiation & SBT collected by GOES 11 C3 & 18 \\
                  & SBT collected by GOES 11 C4 & 19 \\
                  & SBT collected by GOES 12 C3 & 4 \\
                  & SBT collected by GOES 12 C4 & 17 \\
        \bottomrule
    \end{tabular}
    \caption*{SBT stands for Surface Brightness Temperature, a satellite-derived radiation variable included under the Radiation group; GOES stands for Geostationary Operational Environmental Satellite; C stands for channel.}
\end{table*}

\section{Methodology}
\subsection{Problem Formulation}
We formalized the precipitation prediction task as follows. Let \( X \) denote the original set of spatio-temporal input features, and let \( Y \) represent the corresponding target precipitation values. The goal is to learn a predictive function \( f \) that maps the inputs \( X \) to the outputs \( Y \):

\begin{equation}
Y = f(X)
\end{equation}

To enhance predictive capacity, we extended the input space by incorporating additional external data sources, denoted as \( Z \). These include physically meaningful variables from radar, satellite, and radiation channels. The resulting augmented dataset is defined as:

\begin{equation}
X' = X \cup Z
\end{equation}

To facilitate effective multimodal fusion, we applied an element-wise summation strategy inspired by previous work in multimodal learning~\cite{tsai2019multimodal,lu2019vilbert}. Each modality was first normalized using min-max scaling to ensure numerical consistency. The normalized feature vectors were then combined via \textit{element-wise summation}, followed by a second round of normalization to preserve scale uniformity in the fused input representation.

This formulation facilitated the integration of heterogeneous climate data into a unified feature space, allowing the model to more effectively capture the spatial and temporal patterns crucial for accurate precipitation forecasting.

\subsection{Mixture of Experts Framework}
Like most MoE models, our proposed Adaptive MoE model comprises two key components: expert training and router training. During expert training, we utilized the entire dataset to train each expert individually, thus fostering the development of unique weight distributions that enable them to specialize in handling specific features or modalities. During the router training phase, the router learned to dynamically allocate incoming information to the most appropriate experts, therefore maximizing the utilization of their specialized skills. This intelligent routing mechanism ensured efficient workload distribution and adaptive responses, ultimately enhancing the model's overall performance.
\subsubsection{Experts Training}

In the expert training phase, we constructed an MoE architecture consisting of 16 independent Multi-Layer Perceptron (MLP) networks. Each expert was trained to specialize in capturing distinct patterns from the input climate data.

To promote diversity and reduce redundancy among experts, we implemented a selective training strategy. During each epoch, only two experts were randomly selected for parameter updates, while the others remained frozen (i.e., their weights were not updated when frozen). By rotating the active experts across epochs, the model encouraged each expert to specialize in different subregions of the input space, leading to more effective and diverse representation learning.

To further enforce diversity and reduce overlap between expert functionalities, we introduce a custom loss function:

\begin{algorithm}[t]
\caption{Selective Expert Training with Diversity Loss}
\label{alg:expert_training}
\begin{algorithmic}[1]
\Require Climate dataset $\mathcal{D}$, number of experts $N = 16$, number of epochs $E$
\State Initialize $N$ independent MLP experts: $\{M_1, M_2, \dots, M_{16}\}$
\For{epoch $= 1$ to $E$}
    \State Randomly select two expert indices $i, j \in \{1, \dots, N\}$, with $i \neq j$
    \For{each mini-batch $(\mathbf{x}, \mathbf{y})$ in $\mathcal{D}$}
        \State Freeze parameters of all experts except $M_i$ and $M_j$
        \State $\hat{\mathbf{y}}_i \gets M_i(\mathbf{x})$
        \State $\hat{\mathbf{y}}_j \gets M_j(\mathbf{x})$
        \State Compute prediction losses:
        \State \hspace{1em} $\mathcal{L}_i \gets \text{Loss}(\hat{\mathbf{y}}_i, \mathbf{y})$
        \State \hspace{1em} $\mathcal{L}_j \gets \text{Loss}(\hat{\mathbf{y}}_j, \mathbf{y})$
        \State Compute diversity loss $\mathcal{L}_{\text{div}}$ between $M_i$ and $M_j$
        \State $\mathcal{L}_{\text{total}} \gets \mathcal{L}_i + \mathcal{L}_j + \lambda \cdot \mathcal{L}_{\text{div}}$
        \State Update parameters of $M_i$ and $M_j$ using gradient descent on $\mathcal{L}_{\text{total}}$
    \EndFor
\EndFor
\State \Return Trained expert models $\{M_1, M_2, \dots, M_{16}\}$
\end{algorithmic}
\end{algorithm}

\begin{equation}
\text{Loss} = L_1 + L_2 - \|W_1 - W_2\|
\end{equation}

Here:
\begin{itemize}[leftmargin=1.5em]
    \item \( L_1 \) and \( L_2 \) represent the individual loss terms (e.g., Mean Squared Error) for the two selected experts,
    \item \( W_1 \) and \( W_2 \) are the weight matrices of those experts,
    \item The term \( -\|W_1 - W_2\| \) serves as a diversity regularizer, encouraging the selected experts to learn dissimilar representations.
\end{itemize}

After training, we preserve the specialized knowledge of the experts by saving their parameters, allowing the model state to be reloaded during the router training phase while keeping the expert weights frozen. This design ensures that each expert’s learned specialization is retained and leveraged during inference without being altered by subsequent optimization steps.

\subsubsection{Router Training}

The goal of router training was to enable the router to learn how much each expert should contribute to the final prediction for a given input. To achieve this, we first loaded the pre-trained weights of all 16 experts and froze them to preserve their specialized knowledge. During this phase, only the router's parameters were updated.

For each input instance \( i \), the router assigned a set of weights \( \{w_{i,1}, w_{i,2}, \dots, w_{i,E}\} \) to the outputs of the \( E \) experts. The final prediction \( \hat{y}_i \) was computed as a weighted sum of the individual expert predictions:
\begin{equation}
\hat{y}_i = \sum_{j=1}^{E} w_{i,j} \cdot \hat{y}_{i,j}
\end{equation}
where \( \hat{y}_{i,j} \) is the prediction from expert \( j \) for instance \( i \), and \( w_{i,j} \in [0,1] \) is the router’s assigned weight for that expert.

\paragraph{Loss Function}
The router was trained to minimize the discrepancy between the weighted prediction \( \hat{y}_i \) and the true target \( y_i \), using Mean Squared Error (MSE) as the loss function:

\begin{equation}
\text{MSE Loss} = \frac{1}{N} \sum_{i=1}^{N} (\hat{y}_i - y_i)^2
\end{equation}

Substituting the expression for \( \hat{y}_i \), the loss becomes:

\begin{equation}
\text{Loss} = \frac{1}{N} \sum_{i=1}^{N} \left( \sum_{j=1}^{E} w_{i,j} \cdot \hat{y}_{i,j} - y_i \right)^2
\end{equation}

\noindent
Where:
\begin{itemize}[leftmargin=1.5em]
    \item \( N \): number of training instances in the batch,
    \item \( E \): number of experts,
    \item \( \hat{y}_{i,j} \): prediction from expert \( j \) for input \( i \),
    \item \( w_{i,j} \): router-assigned weight for expert \( j \),
    \item \( y_i \): true target value for input \( i \),
    \item \( \hat{y}_i \): final weighted prediction for input \( i \).
\end{itemize}

This training strategy encourages the router to dynamically route each input to the most relevant expert(s), while maintaining differentiability and enabling end-to-end optimization during router training.

\begin{algorithm}[t]
\caption{Router Training with Frozen Experts}
\label{alg:router_training}
\begin{algorithmic}[1]
\Require Dataset $\mathcal{D} = \{(\mathbf{x}_i, y_i)\}_{i=1}^N$; pre-trained experts $\{M_1, M_2, \dots, M_E\}$; number of epochs $T$; batch size $B$
\State Freeze all expert parameters $\{\theta_1, \dots, \theta_E\}$
\State Initialize router parameters $\phi$
\For{epoch $= 1$ to $T$}
    \For{each mini-batch $\{(\mathbf{x}_i, y_i)\}_{i=1}^B \subset \mathcal{D}$}
        \For{each instance $i = 1$ to $B$}
            \For{each expert $j = 1$ to $E$}
                \State $\hat{y}_{i,j} \gets M_j(\mathbf{x}_i)$ \Comment{Frozen expert prediction}
            \EndFor
            \State $\mathbf{w}_i \gets \text{Router}_\phi(\mathbf{x}_i)$ \Comment{Expert weights from router}
            \State $\hat{y}_i \gets \sum_{j=1}^{E} w_{i,j} \cdot \hat{y}_{i,j}$ \Comment{Weighted final prediction}
        \EndFor
        \State Compute MSE loss:
        \State \hspace{1.5em} $\mathcal{L} \gets \frac{1}{B} \sum_{i=1}^B (\hat{y}_i - y_i)^2$
        \State Update router parameters $\phi$ via gradient descent on $\mathcal{L}$
    \EndFor
\EndFor
\State \Return Trained router parameters $\phi$
\end{algorithmic}
\end{algorithm}

\begin{figure*}[t]
  \centering
  \includegraphics[width=0.95\textwidth]{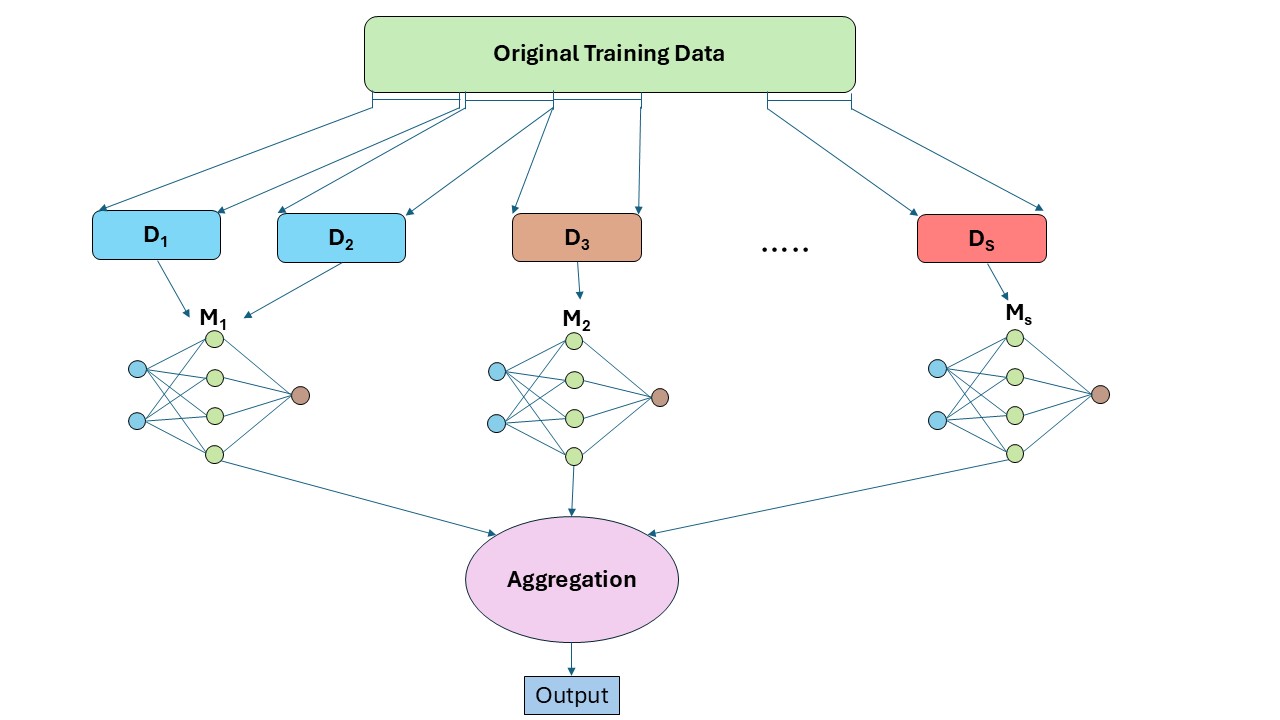}
  \caption{Schematic of the Adaptive MoE model. For conceptual clarity, the training data is depicted as partitioned into multiple subsets ($D_1$ to $D_S$), each associated with a specialized expert model ($M_1$ to $M_S$). In practice, feature grouping was guided by domain knowledge, and data assignments were learned dynamically during pretraining, with inputs routed to the most appropriate experts based on learned weights. The final prediction was obtained by aggregating the outputs of all experts.}
 \label{fig:image2}
\end{figure*}

\section{Experiments Setup}

\subsection{Objectives of the Experiment}

The objectives of this experiment are:
\begin{enumerate}[itemsep=0pt, topsep=0pt, parsep=0pt]
    \item \textbf{Develop an Interactive Web-Based Application:}  
    Create a user-friendly browser-based platform to enable intuitive exploration of the dataset and support feature group visualization.
    
    \item \textbf{Train and Specialize MoE Experts:}  
    Train multiple expert models within an MoE model using the full dataset, encouraging each expert to specialize in distinct feature groups based on domain-informed categorization.
    
    \item \textbf{Evaluate Router-Based Expert Selection:}  
    Assess the effectiveness of a learned routing mechanism that dynamically selects suitable experts for each input instance. This included evaluating the performance gain from adaptive routing compared to non-adaptive or uniformly weighted alternatives.
    
    \item \textbf{Compare Against Baseline Models:}  
    Benchmark the proposed MoE model against conventional baseline models such as MLPs, LSTMs, and Transformer-based models to evaluate improvements in accuracy.
\end{enumerate}

\subsection{Data Preparation}

To support the experimental objectives, the dataset was prepared as follows:

\begin{enumerate}
    \item \textbf{Feature Grouping Based on Domain Knowledge:}  
    The 19 available features were grouped into different categories (e.g., momentum, moisture, radiation) based on domain-specific expertise. This organization enhances both model interpretability and the effectiveness of expert specialization during training.

    \item \textbf{Baseline Setup Using Complete Dataset:}  
    The full dataset, containing all 19 features, was used to establish a performance baseline. The data was divided into training, validation, and test sets using a 7:1:2 ratio. Sequential input samples were generated at 10-day intervals to serve as inputs for the prediction model.
\end{enumerate}

\section{Interactive Web-Based Interface}

\subsection{Design Concepts}

\begin{figure*}[t]
  \centering
  \includegraphics[width=0.75\textwidth]{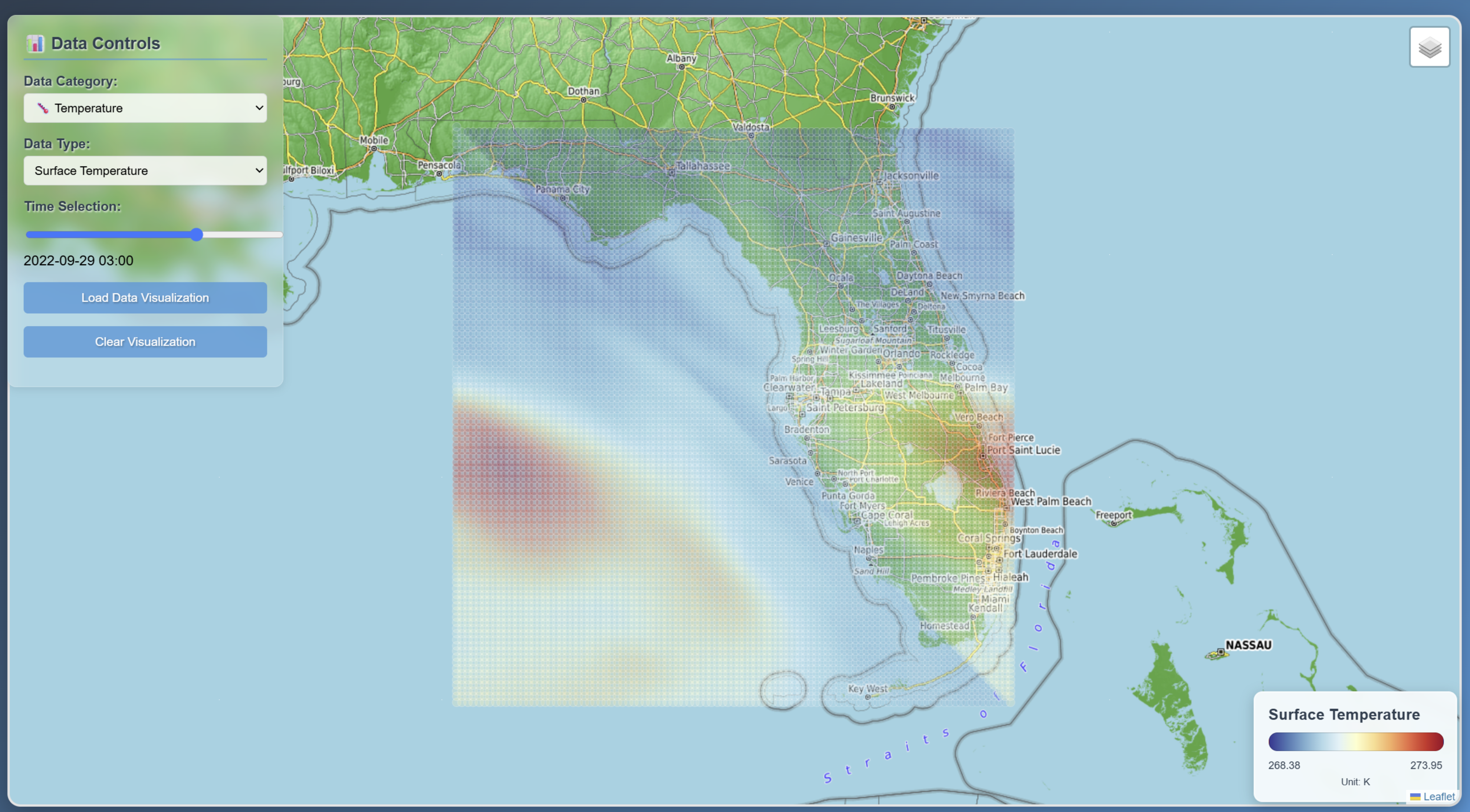}
  \caption{A snapshot of the interactive map from the browser-based website. The current snapshot shows the surface temperature over Florida at 03:00 UTC on September 29, 2022. The gridded overlay visualizes temperature values (in Kelvin), with warmer colors indicating higher temperatures.}
  \label{fig:image3}
\end{figure*}

We developed a lightweight, browser-based interactive interface to facilitate exploration of the climate dataset and knowledge-guided data grouping. An example of the interface is shown in Figure~\ref{fig:image3}. The application is built using Leaflet.js and is geographically constrained to the state of Florida, bounded by latitudes of 24.5°N to 31.0°N and longitudes of 87.0°W to 80.0°W. To minimize visual distractions and maintain consistent spatial alignment across time steps, interactive controls such as panning, zooming, dragging, and keyboard navigation are intentionally disabled. This fixed geospatial framing enables users to more easily track temporal variations without experiencing spatial disorientation.

The interface supports three base map layers:
\begin{itemize}
    \item \textbf{OpenStreetMap:} Highlights road networks and urban infrastructure.
    \item \textbf{Satellite Imagery:} Displays land cover and vegetation details.
    \item \textbf{OpenTopoMap:} Shows topographic contours and elevation.
\end{itemize}

A \emph{glassmorphism} design style is used for control panels, featuring backdrop blur and partial transparency. This approach maintains visibility of the underlying map, offering a visually coherent and aesthetically pleasing experience across different devices and lighting conditions.

\subsection{Data Visualization}

Meteorological variables are organized into seven physically meaningful categories: \textit{Temperature, Hydrology, Radiation, Moisture, Cloud, Mass}, and \textit{Momentum}. Each variable is displayed with its corresponding SI unit and a descriptive label to support accurate scientific interpretation.

Temporal navigation is enabled through an interactive slider, which automatically converts raw filenames (e.g., \texttt{2022-01-01\_08:00}) into clean, human-readable timestamps in the \texttt{YYYY-MM-DD HH:MM} format. The map visualization overlays a uniform spatial grid across the Florida region, with data values represented as colored circles positioned at each grid intersection.

To ensure visual clarity and consistency across all variable types and scales, we applied a \textit{perceptually uniform colormap}. The color gradient ranges from deep blue (low values) to dark red (high values), transitioning through white, yellow, and orange to highlight intermediate levels.

Interactive pop-up windows appear when users hover over or click on a grid point. These windows display detailed metadata, including:
\begin{itemize}
    \item Geographic coordinates and grid index,
    \item Timestamp and exact variable value with unit,
    \item Variable name and standardized description.
\end{itemize}

\subsection{Responsiveness and Usability}

The interface is fully mobile-responsive, allowing field researchers and practitioners to access visualizations in real time from handheld devices. Data is loaded asynchronously, accompanied by an animated progress bar to indicate activity, while informative error messages are displayed when requests fail. A dynamic legend automatically updates to reflect the currently selected variable, adjusting both the color scale and units to ensure consistent and clear interpretation.

This tool not only enhances the interpretability of model predictions but also serves as a practical decision-support platform for stakeholders in agriculture, disaster response, and environmental monitoring. Representative screenshots of the interface are provided in the accompanying figures.

\section{Baselines and Evaluation Metrics}

We employed several baseline models to assess the effectiveness of our proposed method. These included an MLP with three hidden layers and ReLU activations, an LSTM~\cite{hochreiter1997long} network with two stacked LSTM layers, a conventional MoE model without specialized gating or domain-specific adaptations, and a transformer-based model~\cite{vaswani2017attention}. These models were selected to represent a range of architectures commonly employed in sequential and multimodal learning tasks.

To evaluate model performance, we used three standard regression metrics: Mean Absolute Error (MAE), Mean Squared Error (MSE), and Root Mean Squared Error (RMSE). The formulas are defined as follows. We did not include the coefficient of determination ($R^2$), as it can be unreliable in spatial-temporal forecasting tasks with low variance or skewed target distributions, often resulting in inflated or negative values that do not meaningfully reflect predictive accuracy.

\begin{equation}
\text{MAE} = \frac{1}{n} \sum_{i=1}^{n} |y_i - \hat{y}_i|
\end{equation}

\begin{equation}
\text{MSE} = \frac{1}{n} \sum_{i=1}^{n} (y_i - \hat{y}_i)^2
\end{equation}

\begin{equation}
\text{RMSE} = \sqrt{\text{MSE}} = \sqrt{ \frac{1}{n} \sum_{i=1}^{n} (y_i - \hat{y}_i)^2 }
\end{equation}

where \( y_i \) denotes the ground truth, \( \hat{y}_i \) denotes the predicted value, and \( n \) is the total number of samples.

\section{Results}

Table~\ref{tab:comparison} presents the performance comparison of various baseline models and our proposed Adaptive MoE model using three standard evaluation metrics: MAE, MSE, and RMSE (mean ± standard deviation). As shown, the MLP yields the highest error across all metrics, while recurrent and attention-based architectures (LSTM and Transformer) achieve better performance than the MLP. The MoE model without pretraining outperforms all other baselines.

Our proposed Adaptive MoE model achieves the best performance across all metrics, with a substantial reduction in both MAE and RMSE, indicating its ability to capture complex patterns in multimodal data.

\begin{table}[h]
\centering

\label{tab:results}
\small
\caption{Performance comparison of baseline and proposed models (mean ± standard deviation). The results are shown in three standard regression metrics: Mean Absolute Error (MAE), Mean Squared Error (MSE), and Root Mean Squared Error (RMSE).}
\label{tab:comparison}
\begin{tabular}{@{}lccc@{}}
\toprule
\textbf{Model} & \textbf{MAE} & \textbf{MSE} & \textbf{RMSE} \\
\midrule
MLP & 0.587 (± 0.015) & 1.186 (± 0.042) & 1.089 (± 0.021) \\
LSTM & 0.400 (± 0.012) & 0.265 (± 0.008) & 0.515 (± 0.010) \\
Transformer & 0.411 (± 0.010) & 0.266 (± 0.009) & 0.516 (± 0.011) \\
MoE (no pretraining) & 0.309 (± 0.008) & 0.140 (± 0.006) & 0.374 (± 0.007) \\
\textbf{Adaptive MoE (ours)} & \textbf{0.212 (± 0.005)} & \textbf{0.057 (± 0.002)} & \textbf{0.238 (± 0.003)} \\
\bottomrule
\end{tabular}
\end{table}

To further investigate the contribution of different model components, we conducted an ablation study (Table~\ref{tab:ablation}). Removing either pretraining or expert specialization results in a noticeable performance drop, with the absence of expert specialization having the most adverse impact. This supports the design choice of using adaptive expert allocation in our MoE architecture.

\begin{table}[h]
\centering
\caption{Ablation study results (mean ± standard deviation) for different model configurations. Performance is evaluated using Mean Absolute Error (MAE), Mean Squared Error (MSE), and Root Mean Squared Error (RMSE).}
\label{tab:ablation}
\small
\setlength{\tabcolsep}{4pt}
\renewcommand{\arraystretch}{1.2}
\begin{tabular}{@{}lccc@{}}
\toprule
\textbf{Configuration} & \textbf{MAE} & \textbf{MSE} & \textbf{RMSE} \\
\midrule
No Specialization & 0.448 (± 0.013) & 0.257 (± 0.010) & 0.507 (± 0.011) \\
No Pretraining & 0.309 (± 0.008) & 0.140 (± 0.006) & 0.374 (± 0.007) \\
\textbf{Adaptive MoE (ours)} & \textbf{0.212 (± 0.005)} & \textbf{0.057 (± 0.002)} & \textbf{0.238 (± 0.003)} \\
\bottomrule
\end{tabular}
\end{table}

\section{Conclusion}
This work presents the knowledge-guided MoE-Climate dataset and an adaptive MoE model for precipitation forecasting using multimodal climate data. Our experimental results demonstrate that the dataset is both viable and informative, capturing diverse meteorological signals across multiple modalities. Feature analysis indicates that predictive power is distributed rather than dominated by any single input type, reinforcing the value of multimodal learning.

The proposed Adaptive MoE model demonstrates strong performance compared to baseline methods. By dynamically routing inputs to specialized experts, our approach maintains robustness and interoperability. Additionally, we developed a lightweight, browser-based interface to support the interactive exploration of model outputs, thereby improving accessibility for researchers and practitioners in climate-sensitive domains. Together, the dataset, model, and visualization tool form a comprehensive resource for advancing research in spatio-temporal climate prediction.

\section{Future Work}

This study opens several promising directions for future work. First, we plan to integrate uncertainty quantification into the Adaptive MoE model to enhance decision-making in high-stakes environmental forecasting scenarios, where confidence in predictions is crucial. Second, we aim to improve the model’s ability to handle missing modalities and incomplete key features, potentially through attention-based gating mechanisms or knowledge distillation from more complete data representations. Third, we intend to expand the dataset to include temporal dynamics from a broader range of climate zones and to incorporate physical constraints into the model architecture to enhance generalizability and interpretability. Additionally, we plan to enhance the current web-based application by adding support for model prediction visualization and evaluation, providing users with a more comprehensive and interactive tool for climate analysis. Finally, we aim to deploy the model in real-time forecasting systems and evaluate its performance. 

\section*{Acknowledgement}
The authors thank Drs. Zijie Zhang and Naiqing Pan for the insightful discussions that inspired this project. We also acknowledge NOAA for providing the climate data used in this study.

\bibliographystyle{unsrt}
\bibliography{main} 

\end{document}